\documentclass[10pt,twocolumn,letterpaper]{article}

\usepackage[pagenumbers]{cvpr} 

\usepackage{url}
\usepackage{multirow}
\usepackage{xspace}
\usepackage{graphicx}
\usepackage{microtype}
\usepackage{amsmath}
\usepackage{amssymb}
\usepackage{mathtools}
\usepackage{amsthm}
\usepackage{booktabs}
\usepackage{color}
\usepackage{colortbl}
\definecolor{Gray}{gray}{0.95}

\usepackage[misc]{ifsym} 

\usepackage[mathscr]{euscript}

\def\bF{{\bf F}}

\def\bY{{\bf Y}}

\def\bH{{\bf H}}
\def\bX{{\bf X}}
\def\matR{{\mathbb{R}}}

\def\OURS{{WiseAD}\xspace}

\definecolor{cvprblue}{rgb}{0.21,0.49,0.74}
\usepackage[pagebackref,breaklinks,colorlinks,allcolors=cvprblue]{hyperref}

\title{WiseAD: Knowledge Augmented End-to-End Autonomous Driving with Vision-Language Model}

\author{%
Songyan Zhang$^{1}$\thanks{Co-first authors}, ~ Wenhui Huang$ ^{1*}$, ~ Zihui Gao$^2 $, ~ Hao Chen$ ^2 $, ~ Chen Lv$ ^{1}\thanks{Corresponding author}$
\\[0.24cm]
$ ^1 $ Nanyang Technology University, Singapore
~~~
$ ^2 $ Zhejiang University, China
}

\begin{document}
\maketitle


\begin{figure*}[h]
    \centering
    \includegraphics[width=0.9\linewidth]{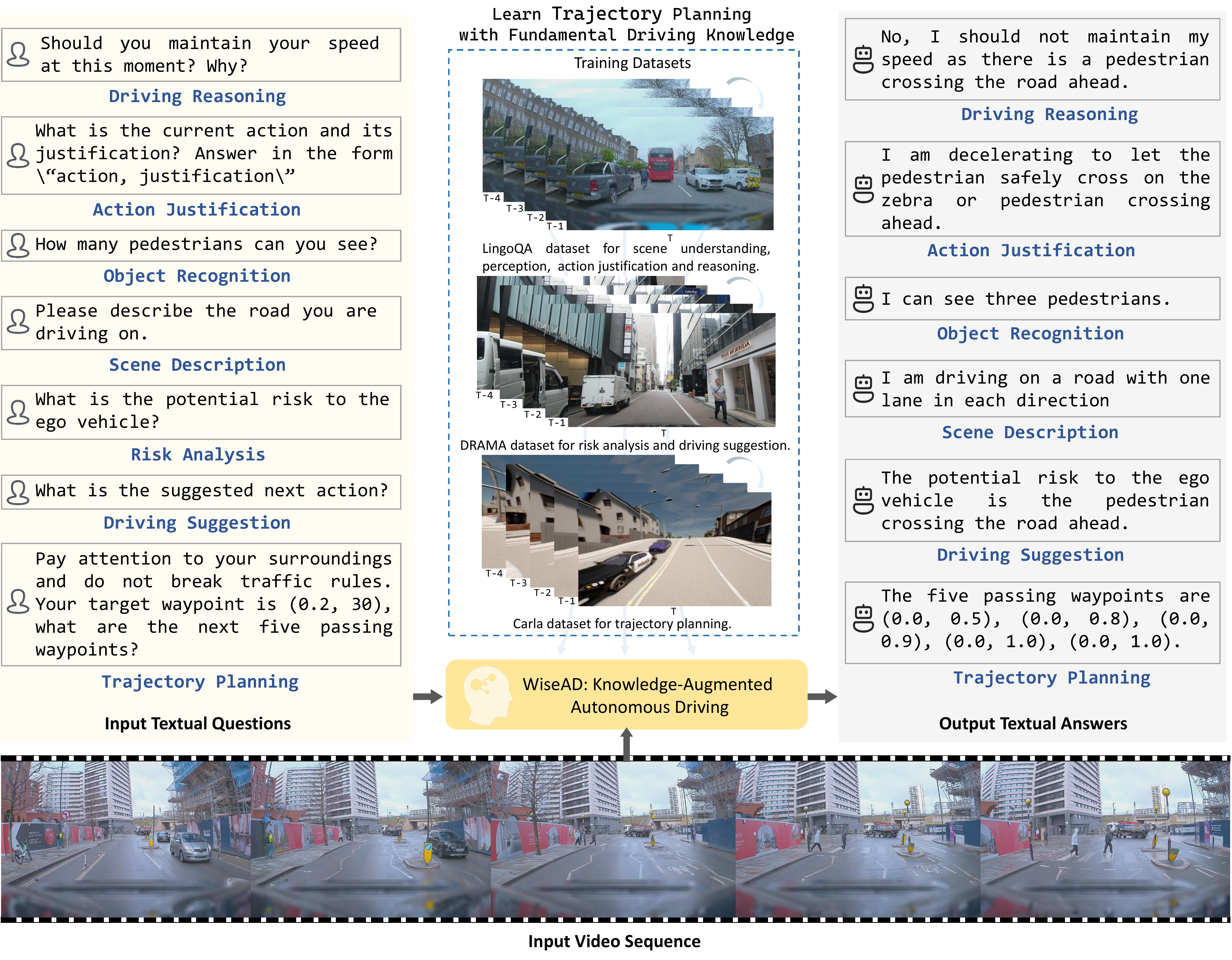}
    \captionsetup{type=figure}
    \caption{\textbf{An overview of the proposed \OURS}, a specialized vision-language model for end-to-end autonomous driving with extensive fundamental driving knowledge. Given a clip of the video sequence, our \OURS is capable of answering various driving-related questions and performing knowledge-augmented trajectory planning according to the target waypoint.
    }
    \vspace{8pt}
    \label{fig:teaser}
\end{figure*}

\begin{abstract}
The emergence of general human knowledge and impressive logical reasoning capacity in rapidly progressed vision-language models (VLMs) have driven increasing interest in applying VLMs to high-level autonomous driving tasks, such as scene understanding and decision-making. However, an in-depth study on the relationship between knowledge proficiency—especially essential driving expertise—and closed-loop autonomous driving performance requires further exploration. In this paper, we investigate the effects of the depth and breadth of fundamental driving knowledge on closed-loop trajectory planning and introduce WiseAD, a specialized VLM tailored for end-to-end autonomous driving capable of driving reasoning, action justification, object recognition, risk analysis, driving suggestions, and trajectory planning across diverse scenarios. We employ joint training on driving knowledge and planning datasets, enabling the model to perform knowledge-aligned trajectory planning accordingly. Extensive experiments indicate that as the diversity of driving knowledge extends, critical accidents are notably reduced, contributing 11.9\% and 12.4\% improvements in the driving score and route completion on the Carla closed-loop evaluations, achieving state-of-the-art performance. Moreover, WiseAD also demonstrates remarkable performance in knowledge evaluations on both in-domain and out-of-domain datasets.
\end{abstract}    
\section{Introduction}
With the advancements of related modules including perception, prediction, planning, control, $etc$, autonomous driving has made significant progress in recent years, transitioning from the traditional rule-based system \citep{ap} to the end-to-end solution\citep{uniad, vad, stp3}. Despite the impressive breakthroughs achieved across various benchmarks, autonomous driving still faces challenges in scene understanding and struggles to leverage fundamental driving knowledge for reliable trajectory planning as a mature human driver, which may potentially hinder further development. 

Recently, the emergent general intelligence exhibited by vision-language models (VLMs) \citep{qwen-chat, internvl, deepseek-vl, llava, mobilevlm-v1, mobilevlm-v2} has demonstrated a remarkable ability to comprehend visual content and perform sophisticated vision-language dialogues based on the visual and textual inputs. This suggests a potential solution for enhancing autonomous driving to emulate human drivers more closely. However, exploiting this general intelligence and harnessing the logical reasoning capabilities for trustworthy trajectory planning is non-trivial. The primary challenge is twofold: (1) \textit{Shortage of driving-oriented knowledge in VLMs.} Following \citep{knowledge}, we refer to the concretization and generalization of human representation of driving scenes, driving experiences, and causal reasoning as fundamental driving knowledge. The widely used VLMs are primarily designed to develop a broad cognitive understanding of the world. It has been demonstrated \citep{coda, LingoQA} that the direct application of vanilla VLMs to answer driving-related questions leads to redundant and meaningless correspondence. 
(2) \textit{Shortage of knowledge alignment for trajectory planning}. Given a target waypoint as the guidance, the task of trajectory planning is to formulate a reasonable path to reach the destination. 
Although pioneering works \citep{interfuser, reasonnet, uniad, lmdrive, vad} have investigated the integration of various modules such as perception and prediction, and exploited advantages of multi-modal sensor fusion, the learning of navigation focuses on imitating the driving behavior of pre-defined agents while neglecting the essential driving knowledge behind. For example, autonomous vehicles may decelerate and adopt cautious driving behaviors in areas with roadside parking vehicles. However, they still struggle with understanding that these decisions are intended to prevent the sudden appearance of pedestrians, thereby avoiding collisions and ensuring safety, highlighting the need for further exploration of explicit knowledge embedding.

In this paper, we aim to tackle these two challenges by proposing \OURS, a vision-language model tailored for autonomous driving with extensive fundamental driving knowledge covering scene understanding, object recognition, potential risk analysis, driving action reasoning, driving action suggestion, and is capable of planning trajectory according to the learned knowledge. An overview of our \OURS is illustrated in Fig \ref{fig:teaser}. MobileVLM(1.7B) \citep{mobilevlm-v2} is leveraged as our vision-language model which is a lightweight yet efficient framework targeted for mobile scale. To extend the driving-oriented essential knowledge, we first collect video question-answering datasets including LingoQA \citep{LingoQA}, and DRAMA \citep{drama}, expanding both knowledge depth (diverse scenarios) and knowledge width (various tasks). To align trajectory planning with driving expertise, we integrate the knowledge and trajectory planning data for joint learning, which enables the model to learn how to infer the future trajectory and why such a path is planned. Furthermore, to seamlessly incorporate the linguistic capabilities of vision-language models, the representation of planned trajectories is also unified under textual scope as DriveVLM\citep{drivevlm}.

Comprehensive experiments are conducted to demonstrate that our \OURS efficiently enhances trajectory planning with essential driving knowledge, achieving a significant improvement in driving score and route completion along with significantly reduced vital accidents such as collisions and running traffic lights. Besides, as the wisdom extends, question-answering capability about driving obtains an obvious promotion for both in-domain and out-of-domain knowledge evaluation. Our contributions can be summarised as follows:

\begin{itemize}
\itemsep 0cm
    \item We propose \OURS, a knowledge-augmented vision-language model tailored for autonomous driving. This model incorporates extensive foundational driving knowledge collected from diverse scenarios, enhancing general driving-oriented cognition in areas such as driving reasoning, action justification, object recognition, scene description, risk analysis, driving recommendations, and trajectory planning.

    \item Through extensive experiments, we demonstrate that expanding the depth and breadth of knowledge with a rationale training paradigm consistently improves both knowledge evaluation outcomes and end-to-end driving performance.
    
    
    
    \item Our comprehensive experiments demonstrate the effectiveness of our \OURS on both closed-loop driving and driving-related knowledge evaluations, achieving state-of-the-art performance.
\end{itemize}
\section{Related Works}
\subsection{LLMs and VLMs for Autonomous Driving}
The pioneering work ADAPT \cite{adapt} made the early exploration to leverage the video swin transformer \cite{video-swin-transformer} for textual driving narration and reasoning, which provides an explicit explanation of driving behavior. DriveGPT4\cite{drivegpt4} shares a similar idea of using VLM for interpretable end-to-end driving with extensive training data. LMDrive \cite{lmdrive} proposes an end-to-end autonomous driving model based on LLaVA \cite{llava} to process multi-modal sensor data with natural language instructions. In DriveVLM \cite{drivevlm}, a slow-fast hybrid system for autonomous driving is proposed where VLM is responsible for scenario understanding and planning enhancement. Another traditional pipeline is also integrated to meet the real-time inference requirement. DriveMLM \cite{drivemlm} incorporates additional lidar data and proposes a multi-modal model based on LLaMA \cite{llama} to provide high-level driving decisions. Instead of providing an end-to-end solution, RAG-Driver \cite{RAG} uses the VLM for knowledge retrieval and enhanced generalizable driving explanations. ELM \cite{elm} integrates multiple driving tasks like object detection, activity prediction, tracking, and scenario description. The limitation of ELM is that the proposed VLM agent couldn't provide either future trajectories or driving decisions which hinders further exploration of closed-loop driving performance. 

\subsection{Knowledge-Augmented Dataset for Autonomous Driving}
As discussed in \citep{knowledge-driven}, autonomous driving is gradually evolving into knowledge-driven technologies, which is greatly attributed to the emergence of knowledge-augmented datasets. Compared with traditional driving datasets \citep{KITTI, cityscapes, nuscenes, waymo} with standard annotations for perception and other tasks, knowledge-augmented datasets normally introduce textual captions for explicit expertise expression. BDD-X \citep{bddx} is proposed for trustworthy and user-friendly autonomous driving. It is composed of over 77 hours of videos with additional textual justifications for driving actions and has been widely used for evaluating vehicle control, explanation generation, and scene captioning. HAD dataset \citep{HAD} is collected from HDD dataset \citep{HDD} and contains 5675 driving video clips to provide explicit driving advice annotated by humans. The driving suggestion covers speed, traffic conditions, road elements, and driving maneuvers. Safety has always been a critical challenge for autonomous driving, and DRAMA dataset \citep{drama} aims to provide explicit risk analysis in terms of object and scenario level with accompanying textual driving suggestions. Recently, the proposal of CODA-LM \citep{coda-lm} dataset collects various long-tail corner cases and provides textual annotations for general perception, regional perception, and driving suggestions. NuScenes dataset \citep{nuscenes} is a popular dataset with abundant annotations for perception, prediction, and planning tasks and has been broadly adopted in traditional solutions. Recently, some explorations have been conducted to provide NuScenes dataset with textual knowledge annotations. DriveLM \citep{drivelm} proposes a graph-style structure to connect the question-answering pairs across perception, prediction, and planning tasks. Instead of using video clips, only keyframes are selected. Talk2Car \citep{talk2car} and NuScenesQA \citep{nuscenesqa} datasets are also built on Nuscenes Dataset while the former dataset focuses on converting driving commands and the latter dataset concentrates on the manual construction of scene graphs and questions by leveraging existing 3D detection annotations. In LingoQA \citep{LingoQA}, authors explore a truthfulness classifier named Lingo-Judge with a higher correlation coefficient to human evaluations. Besides, a comprehensive video question-answering dataset is proposed including tasks of driving reasoning, object recognition, action justification, and scene description. The introduction of CoVLA dataset \citep{covla} encompasses 10,000 video clips incorporating language captions describing the driving scenarios as well as the future trajectory actions. In this work, we focus on the video-based question-answering data pairs and leverage LingoQA, DriveLM, DRAMA datasets for learning fundamental driving knowledge.
\section{Methodolgy}
\subsection{Overview of \OURS}
Our proposed \OURS is a specialized vision-language model with extensive fundamental driving knowledge tailored for autonomous driving, capable of scene description, object recognition, action justification, potential risk analysis, driving suggestions, and trajectory planning. The output is aligned to textual space as DriveVLM\cite{drivevlm} so that the linguistic capability from the pre-trained model can be well preserved.

\begin{figure*}[!ht]
\centering
\includegraphics[width=1\textwidth]{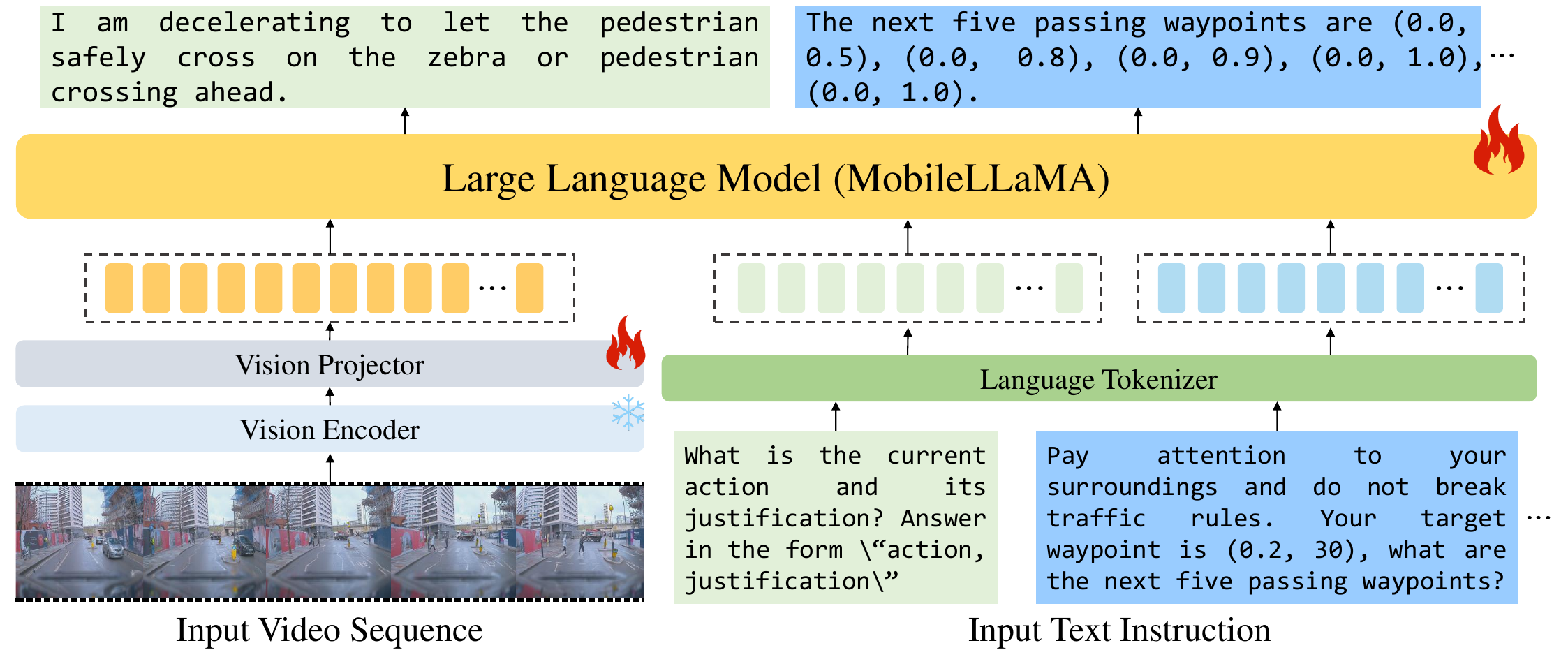}
\caption{\textbf{The framework of the \OURS}. Our model is built upon the MobileVLM and takes video sequences and textual prompts as input. The output for corresponding answers is unified into the linguistic expression to leverage the logical reasoning capability in vision-language models.}
\label{fig:WiseAD}
\end{figure*}

Our proposed \OURS is built upon MobileVLM\citep{mobilevlm-v2}, a computation-friendly vision-language model targeted for mobile devices. The overall framework is illustrated in Fig.\ref{fig:WiseAD}, consisting of a frozen CLIP ViT-L/14 \citep{clipvit} with a learnable projector for visual tokens extraction and a large language model MobileLLaMA for textual questioning and answering. Particularly, given a video sequence of $T$ images $\bX_v \in \matR^{T \times H \times W \times 3}$, the CLIP ViT features $\bF_v \in \matR^{T \times N_v \times D_v}$ are projected to modality-aligned visual tokens $\bH_v \in \matR ^{T \times \frac{N_v}{4} \times D_l}$ where $D_v$ and $D_l$ denote hidden dimension of ViT and MobileLLaMA embeddings and $N_v=HW/14^2$. The projected visual tokens are then flattened along the temporal dimension. The language prompt $\bX_l$ is tokenized to text tokens $\bH_l \in \matR^{N_l \times D_l}$ following the concatenation with $\bH_v$, where $N_l$ is the textual sequence length. The large language model takes the multimodal tokens and generates the corresponding textual response $\bY_a$ of length $L$ via autoregression: 
\begin{align}
    p(\bY_a|\bH_v, \bH_l) = \prod_{i=1}^{L}{p(y_i|\bH_v, \bH_l, y<i)},
\end{align}

where $p(\bY_a$) is the probability of the target answers $\bY_a$. For closed-loop driving inference, the generated textual waypoints are converted to the numerical format. Two PID controllers are employed to adjust the steer, throttle, and braking for tracking the heading and velocity as LMDrive \citep{lmdrive}.

\subsection{Data Construction}\label{sec:data_collection}
High-quality data plays a critical role in training vision-language models. In this subsection, we will discuss the formulation of training data for fundamental driving knowledge and trajectory planning. 

\textbf{Fundamental Driving Knowledge:} A mature and trustworthy human driver makes reliable decisions based on accumulated historical information. To emulate this, we collect video-based datasets including LingoQA \citep{LingoQA}, DRAMA \citep{drama} for knowledge learning, and use BDDX \citep{bddx}, DriveLM \cite{drivelm}, and HAD \cite{HAD} datasets for knowledge evaluation. For the LingoQA dataset, we follow the default configuration, where each data pair consists of 5 consecutive frames accompanied by questions and answers about driving reasoning, action justification, object recognition, and scene description. We split the original DRAMA dataset into two sets of driving suggestions and potential risk analysis to explore the effectiveness of introducing additional knowledge domains and scenarios. In DRAMA, BDDX, and HAD datasets, the original video sequences are segmented into 5 frames with evenly spaced sampling intervals. We reformulate questions using fixed question templates based on the original textual descriptions. For the DRAMA dataset, questions are constructed as "\textit{What is the potential risk in the current scenario?}" and "\textit{What is the suggested next action?}". For the BDDX dataset, the question is "\textit{What is the action of the ego car?}". The corresponding answers remain unchanged as the default description in the original datasets. Fixed question template "\textit{What the driver should pay attention?}" is used for the HAD dataset to reflect the knowledge acquisition of driving attention, which is a close task to potential risk analysis. DriveLM \cite{drivelm} dataset was constructed on the keyframe. We sample data pairs for the object recognition task and incorporate the previous 4 frames of the current timestep. 

\begin{table*}[ht]
    \centering
    \setlength\tabcolsep{1pt}
    \renewcommand\arraystretch{1.2}
    \small
    \begin{tabular}{c|c|c|c|c} \hline
     Dataset & Task & Question Form & Answer Form & Num \\ \hline
     \small LingoQA$^*$ & \small \begin{tabular}{@{}c@{}}Driving Reas., Object Recog., \\ Action Just., Scene Desc. \end{tabular} & 
     \small \begin{tabular}{@{}c@{}}Should you decrease your \\ acceleration? State your reason. \end{tabular} & \small \begin{tabular}{@{}c@{}}Yes I should decelerate  \\ because the traffic light is red. \end{tabular} & 41.4w \\ \hline
     
     \small DRAMA & \small Risk Anal., Driving Sugg. &
     \small \begin{tabular}{@{}c@{}}What is the potential risk? \\ What is the suggested next action? \end{tabular} 
     & \small \begin{tabular}{@{}c@{}} There is pedestrian crossing the road.
     \\ Maintain the current driving behavior. \end{tabular} & 3.5w \\ \hline
     
     \small Carla & \small Trajectory Planning & 
     \small \begin{tabular}{@{}c@{}} Your target point is ($x,y$). What are
     \\  the next five passing waypoints? \end{tabular} & 
     \begin{tabular}{@{}c@{}} 
     The next five passing waypoints \\ are ($x_0, y_0$), ($x_1, y_1$) ... \end{tabular}
     & 28w \\ \hline
     
     \small BDDX & Action Rec. & What is the action of ego car? & The car slows down. & 500 \\ \hline
     
     \small DriveLM$^*$ & Object Rec. & What are objects to the front of the ego car? & There are many barriers, ... & 500 \\ \hline
     
     \small HAD & Driving Attention. & What the driver should pay attention? & There are crossing cyclists in the driving lane. & 500 \\ \hline
    \end{tabular}
    \caption{Detailed data construction for training and evaluating knowledge and trajectory planning. * indicates the questions are not fixed.}
    \vspace{1mm}    \label{tab:data_construction}

\end{table*}
\vspace{-1mm}

\textbf{Textual Trajectory Planning:} Following pioneering works \citep{lmdrive,interfuser}, we use the Carla simulator \citep{carla} to collect trajectories of an autopilot running across various scenarios at a constant frequency of approximately 10Hz. Trajectory planning for the next five waypoints is learned based on five adjacent frames from the first view, along with a destination waypoint that specifies the latitudinal and longitudinal distance from the ego vehicle. At the training stage, the target waypoint is expressed as "\textit{Your target waypoint is (x, y), what are the next five passing waypoints?}". The sign of $x$ indicates the steering direction along the horizontal axis, where a positive value represents a right turn and a negative value represents a left turn. The corresponding answer is structured as "\textit{The next five passing waypoints are (x1, y1), (x2, y2), (x3, y3), (x4, y4), (x5, y5).}".

At the inference stage, an attention prefix prompt is introduced as \textit{"Pay attention to your surroundings and do not violate traffic rules. Your target waypoint is (x, y), what are the next five passing waypoints?"}. This attention prefix serves as a trigger to leverage the learned knowledge, facilitating knowledge-augmented trajectory planning and contributing to a notable reduction in accidents. More details will be discussed in Sec \ref{subsec:attention-prefix-prompt}.

\subsection{Joint Learning for Knowledge-Augmented Trajectory Planning.}
Leveraging extensive fundamental driving knowledge to enhance trajectory planning is a non-trivial challenge. An intuitive approach is to leverage the fundamental knowledge data for pre-training at the first stage, followed by the finetuning on the trajectory data at the second stage. However, we observed this two-stage sequential training leads to significant forgetting of driving expertise and a degeneration in navigation performance. Inspired by the human learning process where versatile intelligence and a general logical reasoning capacity contribute to the foundation of learning, we emphasize the importance of loading parameters pre-trained on large-scale data. Furthermore, during the learning process of driving, learners alternate between acquiring theoretical knowledge and applying it through practical experience. In alignment with this process, we propose to jointly learn theoretical knowledge and trajectory planning by mixing the data of these two tasks in approximately equal proportions. Given a batch of training data, the vision-language model is asked to answer driving-related questions across various tasks while simultaneously generating a reliable route to the destination. This joint learning manner facilitates the understanding of essential knowledge underlying driving behaviors. As the diversity of fundamental knowledge accumulates, there is a notable reduction in critical accidents like vehicle collisions, coupled with an increased ratio of route completion. More details about experimental results will be discussed in Sec \ref{sec:exp}.

\section{Experiment}\label{sec:exp}

\begin{table*}[t]
    \centering
    \setlength\tabcolsep{3pt}
    \renewcommand\arraystretch{1.2}
    \begin{tabular}{l|c|c|c|c|c|c|c|c} \hline
    \multirow{3}{*}{Datasets} & \multirow{3}{*}{\begin{tabular}{@{}c@{}}Pretrained \\ model \end{tabular}} 
    & \multicolumn{6}{c|}{Carla Closed-Loop Eval} & \multirow{3}{*}{\begin{tabular}{@{}c@{}}LingoQA \\ Eval$\uparrow$ \end{tabular}} \\ \cline{3-8}
    
    & & \begin{tabular}{@{}c@{}}Driving \\ score$\uparrow$ \end{tabular} 
    & \begin{tabular}{@{}c@{}}Route \\ compl$\uparrow$ \end{tabular} 
    & \begin{tabular}{@{}c@{}}Infrac. \\ score$\uparrow$ \end{tabular} 
    & \begin{tabular}{@{}c@{}}Red light \\ infraction$\downarrow$ \end{tabular}  
    & \begin{tabular}{@{}c@{}}Collision \\ vehicle$\downarrow$ \end{tabular} 
    & \begin{tabular}{@{}c@{}}Agent \\ blocked$\downarrow$ \end{tabular}\\ \hline
    \multicolumn{9}{c}{Sequential Training} \\ \hline
    Carla & MobileVLM & 62.46 & 83.47 & 0.74 & 2.60 & 2.35 & 0.14 & 13.4 \\ \hline
    Carla & LingoQA-Pre & 52.30 & 75.11 & 0.71 & 2.78 & 2.85 & 3.87 & 12.8 \\ \hline
    \multicolumn{9}{c}{Joint Training} \\ \hline
    Carla+LingoQA & MobileVLM & 63.80 & 87.96 & 0.72 & 3.79 & 5.60 & 0.64 & 58.2 \\ \hline
    \begin{tabular}{@{}l@{}}Carla+LingoQA \\ +DRAMA Suggestion\end{tabular}
     & MobileVLM & 66.02 & 89.50 & 0.75 & 2.26 & 1.87 & 1.37 & 58.4 \\ \hline
    \begin{tabular}{@{}l@{}@{}}Carla+LingoQA \\ +DRAMA Suggestion \\ +DRAMA Risk \end{tabular} & MobileVLM & \textbf{69.88} & \textbf{93.79} & \textbf{0.76} & \textbf{2.14} & \textbf{1.43} & \textbf{0.14} & \textbf{60.4} \\ \hline    
    \end{tabular}
    \caption{Experiment on data diversity and training recipe. The best performance is reported in \textbf{bold}. The increasing depth and width of training data introduces consistent improvement with joint learning of both trajectory planning and driving knowledge.}\label{tab:data_scaling}
    \vspace{1mm}
\end{table*}

\begin{table*}[ht]
    \centering
    \setlength\tabcolsep{6pt}
    \renewcommand\arraystretch{1.2}
    \begin{tabular}{l|c|c|c|c|c|c|c|c|c} \hline
    \multirow{2}{*}{Dataset} & \multicolumn{3}{c|}{LingoQA$\uparrow$} & \multicolumn{2}{c|}{BDDX-Action$\uparrow$} & \multicolumn{2}{c|}{DriveLM-Obj$\uparrow$} & \multicolumn{2}{c}{HAD-Attention$\uparrow$}  \\ \cline{2-10}
    & L-Judge & BLEU & CIDEr & BLEU & METEOR & BLEU & METEOR & BLEU & METEOR  \\ \hline
    C & 13.4 & 2.2 & 21.5 & 0.0 & 2.2 & 4.4 & 4.0 & 0.0 & 0.2 \\ \hline
    C+L & 58.2 & 13.7 & 64.3 & 0.3 & 3.5 & 16.9 & 14.9 & 0.8 & 5.9 \\ \hline 
    C+L+D & \textbf{60.4} & \textbf{14.2} & \textbf{68.3} & \textbf{1.5} & \textbf{10.8} & \textbf{16.9} & \textbf{15.5} & \textbf{0.9} & \textbf{6.1} \\ \hline

    \end{tabular}
    \caption{Experiment on the effectiveness of extending training data on knowledge evaluation. C, L, D are short for Carla, LingoQA, DRAMA datasets, respectively. The best performance is reported in \textbf{bold}.}\label{tab:knowledge_ablation}
    \vspace{1mm}
\end{table*}
\vspace{-1mm}

\subsection{Data Analysis}
Our proposed \OURS is trained on a mixture of various datasets including the LingoQA \citep{LingoQA}, DRAMA \citep{drama} for learning fundamental driving knowledge, along with learning trajectory planning on the Carla \citep{carla} dataset. The detailed number of data pairs and their corresponding functions for each dataset is illustrated in Tab.\ref{tab:data_construction}. During an epoch of training, the Carla dataset is sampled twice for a balanced proportion of theoretical knowledge and trajectory planning.

We adopt the configuration in LMDrive \citep{lmdrive} for sampling trajectory data with a pre-defined rule-based agent in the Carla simulator.  It's worth mentioning that only first-view images are collected in consistency with the data format in knowledge datasets. During the evaluation stage, the LingoQA validation dataset, BDDX \citep{bddx}, DriveLM \cite{drivelm}, and HAD \cite{HAD} datasets are employed to validate different knowledge acquisition through QA tasks, while trajectory planning is evaluated through zero-shot testing in the Town05 environment of CARLA within a closed-loop sense. We randomly sample 500 data pairs for each of BDD-X, DriveLM, and HAD datasets to construct the zero-shot evaluation for action justification, object recognition, and driving attention, respectively. 

\subsection{Implementation Details}
We leverage the training framework in MobileVLM \citep{mobilevlm-v2} and initialize the model with parameters of instruction tuning. The whole training process on the mixture data takes 2 epochs with the peak learning rate of $4\times 10 ^{-5}$ for the first epoch and $1\times 10 ^ {-5}$ for the second epoch. The cosine strategy is adopted for both epochs accompanied by the warming-up ratio of 0.03 and 0.1, respectively. We use the AdamW \citep{adamw} optimizer and a global batch size of 128 on 4 NVIDIA Tesla A100 (40GB) GPUs.

\subsection{Evaluation Metric}
For driving-related knowledge evaluation, we follow previous works \citep{elm, LingoQA, llava} and report two established metrics of CIDEr \citep{cider}, BLEU \citep{bleu}. Moreover, we leverage the Lingo-Judge in LingoQA which is a pretrained transformer-based text classifier to evaluate the knowledge proficiency on the LingoQA dataset. Given a question, the human's answer, and the vision-language model's prediction, the Lingo-Judge estimates the probability that the model's answer is correct. For closed-loop evaluation on the Carla simulation, we consider three primary metrics including route completion (RC), infraction score (IS), and driving score (DS). The route completion depicts the percentage of the route that has been completed. The infraction score indicates infractions triggered by the agent. The driving score is a comprehensive metric calculated by weighting the route completion and infraction scores. More details can be found in the Carla LeaderBoard \citep{carla}. Additionally, we report the number of routes where crucial accidents of running red lights, and collisions occur to demonstrate the effectiveness of fundamental knowledge regulation.

\begin{table*}[ht]
    \centering
    \vspace{1mm}
    \setlength\tabcolsep{6.5pt}
    \renewcommand\arraystretch{1.2}
    \begin{tabular}{c|c|c|c|c|c|c|c|c|c} \hline
    \multirow{2}{*}{Method} & \multicolumn{3}{c|}{LingoQA$\uparrow$} & \multicolumn{2}{c|}{BDDX-Action$\uparrow$} & \multicolumn{2}{c|}{DriveLM-Obj$\uparrow$} & \multicolumn{2}{c}{HAD-Attention$\uparrow$}  \\ \cline{2-10}
    & L-Judge & BLEU & CIDEr & BLEU & CIDEr & BLEU & CIDEr & BLEU & CIDEr \\ \hline
    LLaVA1.5-7B \citep{llava} & 38.0 & 4.0 & 32.3 & \textbf{2.0} & 7.4 & 7.1 & 8.6 & 0.4 & 0.0 \\ \hline
    MobileVLM-1.7B \citep{mobilevlm-v2}  & 40.0 & 2.8 & 25.2 & 2.1 & 8.8 & 15.6 & \textbf{16.4} & 0.0 & 0.2 \\ \hline
    InternVL2-2B \citep{internvl}& 42.6 & 2.2 & 29.0 & 0.9 & 6.4 & 1.9 & 0.0 & 0.2 & 0.0 \\ \hline
    InternVL2-8B \citep{internvl} & 53.4 & 3.0 & 33.3 & 0.6 & 5.7 & 2.7 & 0.4 & 0.2 & 0.0 \\ \hline
    DeepseekVL-7B \citep{deepseek-vl} & 46.4 & 2.9 & 32.5 & 0.2 & 4.3 & 3.5 & 10.5 & 0.3 & 0.0\\ \hline    
    \rowcolor{Gray}\OURS-1.7B(ours) & \textbf{60.4} & \textbf{19.9} & \textbf{68.3} & 1.5 & \textbf{10.8} & \textbf{16.9} & 15.5 & \textbf{0.9} & \textbf{6.1}\\ \hline
    \end{tabular}
    \caption{Comparison with other state-of-the-art methods on driving knowledge evaluation. The best performance is reported in \textbf{bold}.}\label{tab:driving_knowledge}
\end{table*}
\vspace{-1mm}
\begin{table}[t]
    \centering
    \setlength\tabcolsep{7pt}
    \renewcommand\arraystretch{1.2}
    \begin{tabular}{c|c|c|c} \hline
    Methods & Input View & DS $\uparrow$ & RC $\uparrow$ \\ \hline
    TransFuser \citep{transfuser} & Multiview & 54.52 & 78.41 \\ \hline
    NEAT \citep{neat} & Multiview & 58.70 & 77.32 \\ \hline
    Roach \citep{roach} & BEV & 65.26 & 88.24 \\ \hline
    ST-P3\cite{stp3} & Multiview & 55.14 & 86.74 \\ \hline
    VAD\cite{vad} & Multiview & 64.29 & 87.26 \\ \hline
    \rowcolor{Gray}\OURS(ours) & Firstview & \textbf{69.88} & \textbf{93.79} \\ \hline
    
    \end{tabular}
    \caption{Comparisons with other SOTA methods on the Carla dataset for closed-loop evaluation where our \OURS achieves the best performance. DS is short for driving score and RC is short for route completion. The best performance is reported in \textbf{bold}.}\label{tab:carla_sota}
\end{table}
\vspace{-1mm}

\subsection{Impact of Knowledge Depth and Breadth}\label{subsec:data-and-training}
To investigate the effectiveness of driving knowledge depth and breadth, we first explore the training paradigm and conduct step-by-step incremental experiments. The baseline is learning trajectory planning based on the vanilla MobileVLM model, which is trained on large-scale versatile datasets and thus equipped with general intelligence. We start with the sequential training, first fine-tuning the MobileVLM with the LingoQA dataset and then continually learning trajectory planning. The training process and corresponding closed-loop driving performance are presented in Tab.\ref{tab:data_scaling}. The first two rows indicate that general intelligence offers a reasonable foundation for scene understanding and planning. However, further fine-tuning with the LingoQA dataset results in a notable decline in VLM performance, particularly in closed-loop driving tasks, due to catastrophic forgetting—a common issue in continual learning.

This observation motivates us to shift to a joint learning protocol. More specifically, we simultaneously train fundamental driving knowledge and trajectory planning on the pre-trained MobileVLM weights, enhancing both closed-loop driving performance and knowledge acquisition. As shown in the third row, joint training significantly improves scene understanding, as validated by LingoQA evaluation, and modestly enhances driving performance, specifically in route completion and driving score. However, we observe an increase in undesirable behaviors, such as running red lights and colliding with other vehicles, likely due to extended driving distances—highlighting a need for enhanced traffic rule compliance and safety considerations. To address this, we introduced additional domain knowledge from the DRAMA dataset, focusing on driving suggestions to mitigate these behaviors. Although the improvement in LingoQA performance is minor, traffic rule violations and hazardous behaviors decrease substantially, enhancing all three primary driving metrics. Finally, we achieved further improvements across all metrics by incorporating risk analysis data as additional knowledge on top of prior domains. Compared to training solely on the CARLA dataset, joint learning with foundational driving knowledge improves the driving score by 11.9\%, contributing to a significant drop in all critical driving accidents and, therefore, increasing driving safety.

Moreover, we also explore the effectiveness of extending training data over driving knowledge acquisition. As presented in Tab.\ref{tab:knowledge_ablation}, zero-shot evaluations on the BDDX, DriveLM Object, and HAD Attention datasets witness an increasing improvement in terms of the all the metrics. Overall, expanding knowledge depth (data amount) and breadth (diverse domains) with the rationale training recipe results in consistent enhancements in both knowledge evaluations and end-to-end driving performance.

\subsection{State-of-the-Art Benchmark}

\begin{table}[t]
    \centering
    \vspace{1mm}
    \setlength\tabcolsep{5pt}
    \renewcommand\arraystretch{1.2}
    \label{ablation_exp_architecture}
    \begin{tabular}{c|c|c|c} \hline
    Method
    
    & \begin{tabular}{@{}c@{}}Driving \\ score$\uparrow$ \end{tabular} 
    & \begin{tabular}{@{}c@{}}Route \\ compl$\uparrow$ \end{tabular}
    & \begin{tabular}{@{}c@{}}Infrac. \\ score$\uparrow$ \end{tabular} \\ \hline
    w/o attention prompt & 66.89 & 85.35 & \textbf{0.78} \\ \hline
    w attention prompt & \textbf{69.88} & \textbf{93.79} & 0.76  \\ \hline
    \end{tabular}
    \caption{Ablation studies on the effectiveness of attention-guided prompts. The best performance is reported in \textbf{bold}.}\label{tab:attention_prompts}
\end{table}
\vspace{-1mm}

\begin{figure*}[ht]
\centering
\includegraphics[width=0.86\textwidth]{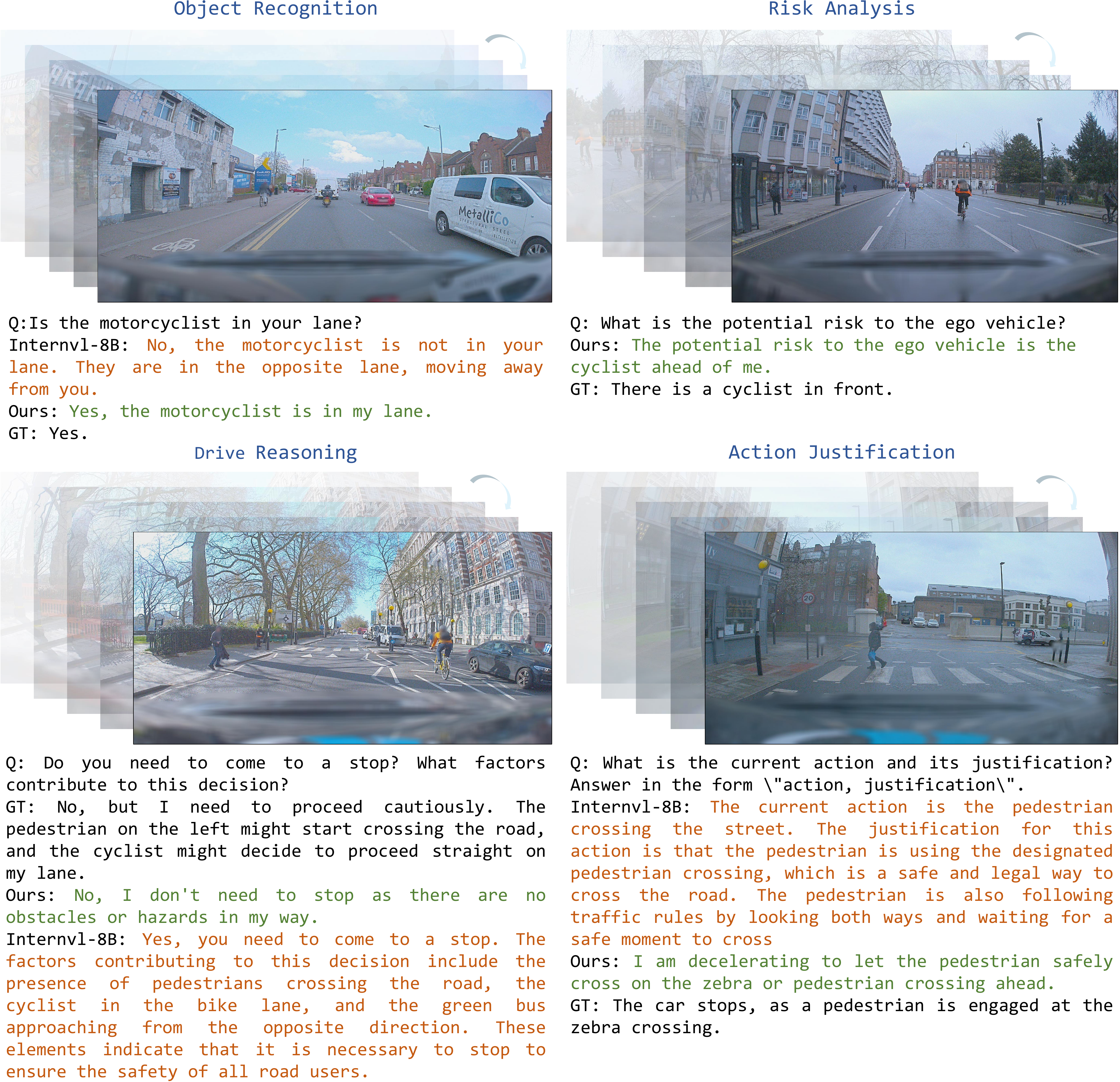}
\caption{Qualitative comparison with the InternVL-8B for driving knowledge evaluation.}
\label{fig:KnowComp}
\end{figure*}

To further validate the positive impact of knowledge augmentation, we compare WiseAD with several representative VLMs equipped only with general knowledge. In addition to the LingoQA test dataset, we employ three additional datasets—BDDX, DriveLM, and HAD—for zero-shot evaluation. As shown in Tab.\ref{tab:driving_knowledge}, WiseAD achieves the highest overall performance on the LingoQA dataset, surpassing other VLMs by a significant margin, even with a smaller model size. Furthermore, its superior performance in zero-shot evaluation demonstrates WiseAD's advantage, providing essential driving priors to execute subsequent closed-loop driving tasks. Finally, qualitative validation results on randomly selected driving scenarios are presented in Fig.\ref{fig:KnowComp}, underscoring the importance of integrating fundamental driving knowledge on top of general intelligence.

As for the closed-loop benchmark, we report the driving performance of zero-shot evaluation on Town05. As shown in Tab.\ref{tab:carla_sota}, with enhanced driving knowledge, our \OURS achieves the best zero-shot driving performance on closed-loop evaluation. Compared with previous representative methods based on data-driven, the metric of route completion is improved by 21\% at most. For the comprehensive metric, the driving score, \OURS outperforms a certain margin against the pioneering end-to-end autonomous driving works VAD \cite{vad} and ST-P3 \cite{stp3}.

\subsection{Ablation Study on Attention-Prefix Prompt}\label{subsec:attention-prefix-prompt}
As discussed in Sec.\ref{sec:data_collection}, we add an attention-based prefix to the target waypoint prompt to explicitly guide the model in leveraging fundamental driving knowledge. As shown in Tab.\ref{tab:attention_prompts}, removing the prefix "Pay attention to your surroundings and do not break traffic rules" leads to a significant performance drop, with route completion decreasing from 93.79 to 85.35 and the driving score declining from 69.88 to 66.89. This result validates that \OURS effectively understands the textual guidance and has the capability to align trajectory planning with learned driving knowledge.
\section{Conclusion}
In this work, we have presented WiseAD, a specialized VLM tailored for end-to-end autonomous driving capable of executing versatile tasks including interactive trajectory planning. WiseAD demonstrates that expanding knowledge depth and breadth with the reasonable training recipe will consistently enhance knowledge evaluations and end-to-end driving performance. In-domain and out-of-domain evaluation results show that closed-loop driving behaviors can be progressively enhanced by injecting previously unlearned domain knowledge. We believe this work provides a reliable foundation for future research focused on closed-loop performance and real-world autonomous driving applications.

{
    \small
    \bibliographystyle{ieeenat_fullname}
    \bibliography{main}
}


\end{document}